\RequirePackage{fix-cm}

\documentclass[times,twocolumn,final]{elsarticle}

\usepackage{ycviu}
\usepackage{framed,multirow}

\usepackage{epstopdf}
\usepackage{import}
\usepackage[normalem]{ulem}
\usepackage{graphicx}
\usepackage{color}
\usepackage{amsmath}
\usepackage{transparent}
\usepackage{subfigure}
\usepackage{algorithm}
\usepackage{amsfonts}
\usepackage{amssymb}
\usepackage{siunitx}
\usepackage[noend]{algpseudocode}

\usepackage{booktabs}
\usepackage{multirow}
\usepackage[table]{xcolor}
\usepackage{tabularx}

\makeatletter
  \def\relativepath{\import@path}
\makeatother

\newcommand{\executeiffilenewer}[3]{%
\ifnum\pdfstrcmp{\pdffilemoddate{#1}}%
{\pdffilemoddate{#2}}>0%
{\immediate\write18{#3}}\fi%
}

\newcommand{\includesvg}[2][]{%
\immediate\write18{inkscape --without-gui --file="\relativepath#2.svg" --export-pdf="\relativepath#2.pdf" --export-area-drawing}%
\ifthenelse{\equal{#1}{}}{%
\includegraphics{#2.pdf}}{%
\includegraphics[#1]{#2.pdf}}%
}

\newcommand*\wrapletters[1]{\wr@pletters#1\@nil}
\def\wr@pletters#1#2\@nil{#1\allowbreak\if&#2&\else\wr@pletters#2\@nil\fi}

\providecommand{\url}[1]{#1}
\let\urlUnwrap\url

\renewcommand*\url[1]{\urlUnwrap{\wrapletters{#1}}}

\journal{Computer Vision and Image Understanding}

\begin{document}
\thispagestyle{empty}

\ifpreprint
\setcounter{page}{1}
\else
\setcounter{page}{1}
\fi

\begin{frontmatter}
	
	\title{3D Reconstruction of Deformable Revolving Object under Heavy Hand Interaction}

	\author[labuom,labinria]{Raoul \snm{de Charette}\corref{cor1}} 
	\cortext[cor1]{Corresponding author.}
	\ead{raoul.de-charette@inria.fr}
	\author[labmines]{Sotiris \snm{Manitsaris}}
	\ead{sotiris.manitsaris@mines-paristech.fr}
	
	\address[labuom]{Multimedia Tech. and Computer Graphics Lab., Uni. of Macedonia, Thessaloniki, Greece}
	\address[labinria]{RITS team, Inria, Paris, France}
	\address[labmines]{Centre for Robotics, Mines ParisTech, France}
	
	\received{1 May 2013}
	\finalform{10 May 2013}
	\accepted{13 May 2013}
	\availableonline{15 May 2013}
	\communicated{S. Sarkar}

\begin{abstract}
We reconstruct 3D deformable object through time, in the context of a live pottery making process where the crafter molds the object.
Because the object suffers from heavy hand interaction, and is being deformed, classical techniques cannot be applied.
We use particle energy optimization to estimate the object profile and benefit of the object radial symmetry to increase the robustness of the reconstruction to both occlusion and noise.
Our method works with an unconstrained scalable setup with one or more depth sensors.
We evaluate on our database (released upon publication) on a per-frame and temporal basis and shows it significantly outperforms state-of-the-art achieving $7.60\text{mm}$ average object reconstruction error. 
Further ablation studies demonstrate the effectiveness of our method.
\end{abstract}

\end{frontmatter}

\section{Introduction}
\label{intro}
\begin{sloppypar}
Human excels at sensing geometry, which allows them to better interact with the environment. 
The reconstruction of 3D objects drew early interests \cite{jain1998deformable} and is of high importance for applications such as virtual reality~\cite{comport2006real} or photogrammetry~\cite{kersten2012potential} (e.g to scan cultural objects).
Still, existing techniques work only in narrow conditions and either require objects known \textit{a priori} or intrusive acquisition setups. Both of which are hardly compatible with complex interactive applications.

\begin{figure}[t]
	\centering
	\subfigure[Pottery making]{\includegraphics[height=6cm]{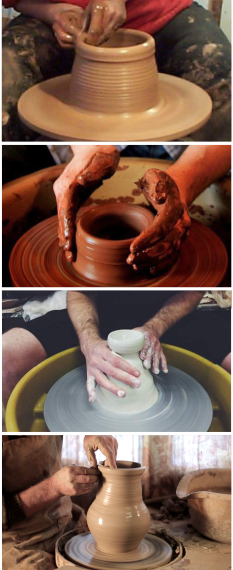}\label{fig:pottery}}
	\hspace{0.06\linewidth}
	\subfigure[Our 3D reconstruction]{\includegraphics[height=6cm]{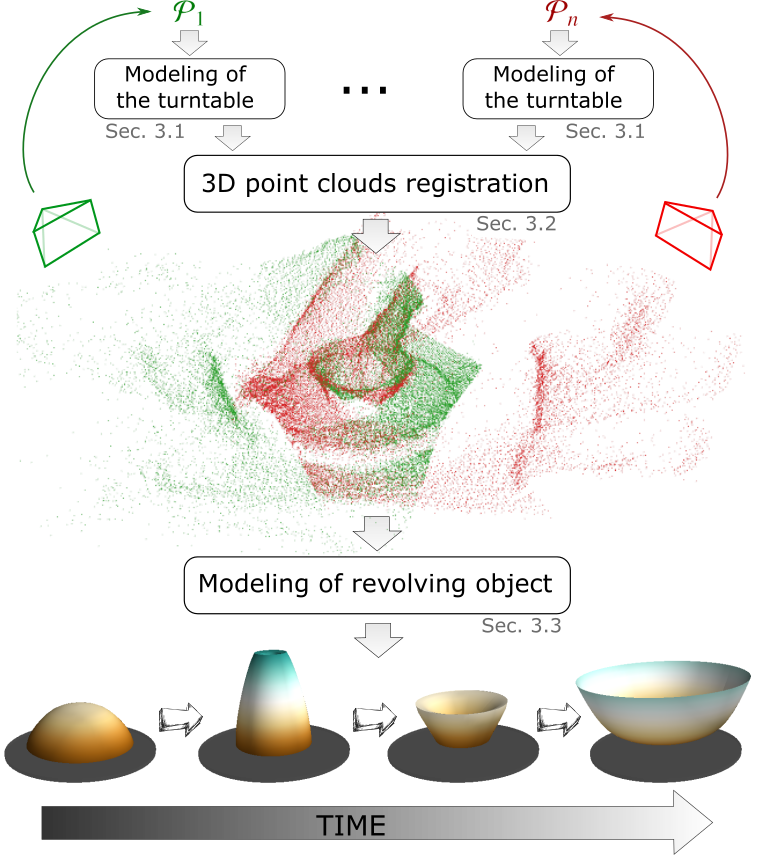}\label{fig:visuMain}}
	\caption{(a) Reconstruction of unknown 3D objects in the context of wheel throwing pottery. (b) Using one or more input point clouds, our methods clusters the 3D scene and extracts the profile of revolving objects. Bottom are sample outputs of our method.}
	\vspace{-2em}
\end{figure}

For photogrammetry purposes, multiple views of the same object (occlusion free) are acquired to compute a 3D model of the object using the photometric and geometric relationship of salient image features~\cite{oikonomidis2011full}. Others model the light-object interaction to estimate the object mesh from shading~\cite{terzopoulos1988constraints}. 
In interactive applications, less views can be acquired simultaneously and with hand interaction the imaging suffers from more occlusion.
In such cases, the common strategy is to assume the object \textit{a priori} known and to track the latter via the energy minimization of textural appearance and geometry \cite{ostlund2012laplacian,tien2015dense,salzmann2009reconstructing,szeliski1993modeling,bartoli2015shape,parashar2015rigid,hilsmann2008tracking}.
When object is unknown, the literature assume rigid objects~\cite{oikonomidis2011full,tsoli2018joint,kyriazis2014scalable,panteleris20153d} or constant-volume object~\cite{parashar2015rigid}.
Such constraints fit the tracking of rigid or articulated objects (paper sheet, doll, robot arm, etc.) but fail when objects evolves through time (clay, sculpture). Likewise photogrammetry requires all-around view of the object, i.e. without occlusion, which is hardly compatible with applications where the object is being manipulated.

\paragraph{Our contribution}
Our method reconstructs arbitrary 3D revolving objects (i.e. radially symmetrical) and copes with deformation and occlusion such as hand interaction. This cannot be addressed with existing techniques. 
The proposal is motivated by the application to wheel throwing pottery where the object shape evolves temporally unpredictably during the modeling phase, see fig.~\ref{fig:pottery}. For this reason, the processing cannot rely on shape or texture priors as the objects exhibit large variety of shapes and clay may cover hand and object indistinguishably. 
Subsequently, we use radially distributed depth sensors, register data together, and estimate the object profile from particle energy minimization. Our setup is unconstrained and the method works with any number of depth sensors.
A composite of our method is shown in fig. \ref{fig:visuMain}. 
\end{sloppypar}

\section{Related work}

\begin{sloppypar}
\paragraph{Known deformable objects}
Reconstruction or tracking of known deformable objects is traditionally coined as Shape from Template (SfT) and assume a known 3D template usually in the form of a 3D textured model~\cite{ostlund2012laplacian,tien2015dense,salzmann2009reconstructing,szeliski1993modeling,bartoli2015shape,parashar2015rigid,hilsmann2008tracking}. 
Its most common application is the tracking of planar surfaces texture (paper, t-shirt, etc.), where the input texture serves as template.
The 3D template may also be extracted using depth only from initial frames~\cite{schulman2013tracking}, or estimated from an image and a 2D silhouette~\cite{vicente2013balloon} using volume inflation techniques~\cite{oswald2012fast}.
SfT is commonly framed as an optimization problem seeking to estimate the template deformer and match the current observation.
It consists of minimizing the topology and appearance energies. 

The topology energy minimizes the geometrical difference with the input models given finite Degrees of Freedom (DoF). 
As opposed to articulated objects, deformable objects may have virtually infinite DoF.
To reduce the complexity, 2d mass springs~\cite{ostlund2012laplacian,tien2015dense,salzmann2009reconstructing,bartoli2015shape} or 3D mass-spring~\cite{schulman2013tracking,vicente2013balloon,parashar2015rigid} can be used. 
In~\cite{parashar2015rigid}, a rigidity energy ensures preservation of the general model shape following the \textit{as-rigid-as-possible} long-time practice in mesh interpolation~\cite{alexa2000rigid} and animation~\cite{sorkine2007rigid}. 

The image-based energy is computed from observation and estimated distortion as the sum of differences of sparse descriptors like SIFT~\cite{lowe2004distinctive}, or dense gradient descriptors~\cite{gopalan2010comparing,crivellaro2014robust}. 
The latter being preferable for poorly texture object~\cite{ostlund2012laplacian,tien2015dense}.

For optimization, the common trend is to use Levenberg-Marquardt or particle swarm. 
For close objects the deformation is computed over planar surfaces~\cite{ostlund2012laplacian,tien2015dense,salzmann2009reconstructing,bartoli2015shape} or object-shell~\cite{schulman2013tracking,vicente2013balloon}. The work of Parashar \textit{et al.}~\cite{parashar2015rigid} proposes a volume-preserving framework leading to finer estimation, though limited to object of constant volume.

All but~\cite{schulman2013tracking} rely heavily on RGB data which isn't compatible with our application as clay covers hand and object.
\end{sloppypar}

\begin{sloppypar}
\paragraph{Unknown objects}
Only a handful of researches explicitly reconstructs free-form unknown object and all seem to assume non-deformable object\footnote{While the object may be of arbitrary topology (e.g. articulated, etc.), its appearance is supposed to be non-evolving during the acquisition.}. 
The early researches~\cite{terzopoulos1988constraints,nastar1993fast,szeliski1993modeling} also required user inputs, such as the object spine~\cite{terzopoulos1988constraints} or coarse geometry~\cite{nastar1993fast} to initialize energy minimization and fit observation, sometimes with additional symmetry criteria~\cite{terzopoulos1988constraints}.

On the other hand, many researches were conducted on unknown scene reconstruction~\cite{newcombe2011kinectfusion,choi2015robust} but are often considered as \textit{not} related to object reconstruction because they require large scene scans and do not model the objects \textit{explicitly}.
However, lines are blurring out with recent data-driven techniques.
For example, OctNetFusion~\cite{riegler2017octnetfusion} predicts an \textit{implicit} object representation from the fusion of truncated signed distance functions~(TSDFs). The extension of~\cite{liao2018deep} even predicts explicit representation from differentiable marching cubes.
Despite good visual results \cite{riegler2017octnetfusion,liao2018deep} they assume outliers-free model, and thus cannot be applied for the present research objective.
\end{sloppypar}

\begin{sloppypar}
\paragraph{Hand-interacting objects}
The tracking and reconstruction of object under strong hand interaction has poorly been addressed and most existing researches were carried out by Argyros and Kyriazis.
Generally, the problem is addressed by simultaneously optimizing an energy for the hand pose and the object model~\cite{oikonomidis2011full,tsoli2018joint,kyriazis2014scalable,panteleris20153d} often using a semi-automatic coarse hand estimate~\cite{tsoli2018joint,panteleris20153d}.
To handle occlusion, ~\cite{oikonomidis2011full} requires 9 cameras but the common strategy is to use a single RGBD input~\cite{tsoli2018joint,kyriazis2014scalable,kyriazis2013physically} or a depth only input~\cite{panteleris20153d}.
To solve ambiguities and reduce complexity, hand-object constraints can also be applied~\cite{kyriazis2013physically,kyriazis2014scalable}. In~\cite{kyriazis2014scalable} an ensemble of collective trackers is used to track hand together with other \textit{rigid} objects. In the literature, the object and the hand are assumed to be visible, except for~\cite{kyriazis2013physically} that uses a physical/collision engine to track plausible hidden motion.
Closer to our research, \cite{panteleris20153d} tracks unknown object from depth data assuming object-fingertip collisions, 
and \cite{Wang2006a} uses BSpline minimization to extract object profile. 
Both lead to great results but require non-deformable object for ICP registration~ \cite{panteleris20153d} or noise-free models~\cite{Wang2006a}.
\end{sloppypar}

\section{Method}
\label{sec:sceneanalysis}

There are several challenges associated with the context of wheel throwing pottery. First, texture is not relevant since hand and pottery are indistinguishably covered with clay~(cf. fig.~\ref{fig:pottery}). Second, the method must be robust to strong occlusion since the potter's models the object during acquisition. 
Note that the wheel cannot be stopped during the making process.
To address these challenges, our setup uses depth sensors radially distributed all around the turntable. 
Our processing lies on two key observations: A)~Pottery object always sits on a turntable, B)~Both pottery object and turntable share a common revolving axis. Overall pipeline is illustrated in fig.~\ref{fig:visuMain}.

For each depth sensor ($i \in \mathbb{N}, i<n$) a 3D point cloud $\mathcal{P}_i$ is acquired. 
Following observation~A, we detect turntable~(sec.~\ref{sec:turntable}) in each point cloud both as it allows further parametric registration and provides us with the revolving axis.
All points clouds are then registered together~(sec.~\ref{sec:PCLRegistration}) using the parametric turntable model.
Observation~B serves to model the pottery object and extract the profile from radial accumulation around the axis of revolution~(sec.~\ref{sec:ModelingRevolvingObject}).

We now detail each step individually.

\paragraph{Notation}
Vectors are with arrow ($\vec{x}$), 
matrices are bold upper letters ($\mathbf{X}$),
 and point clouds are calligraphic letters 
 ($\mathcal{X}$). We use
 $\times{}$~as cross product, 
 $*$~as multiplication, 
 $\cdot{}$~as matrix multiplication, 
 $|.|$~as set cardinality, 
 and $||.||$~as vector/curve length.

\subsection{Modeling of the turntable}
\label{sec:turntable}

The turntable where potters manipulate and sculpt the clay is a plate of unknown position and orientation, parametrized with a center $\vec{c}(x, y, z)$ a normal $\vec{n}(x, y, z)$ and a known radius $r$.
We estimate the full turntable model $(\vec{c}_i, \vec{n}_i, r)$ in each point cloud $\mathcal{P}_i$ from stochastic optimizer and modified kernel density estimator.

\begin{figure}[t]
\centering
	\setlength{\tabcolsep}{0.0\linewidth}
	\renewcommand{\arraystretch}{0.0}
	\subfigure[Normal estimation]{\includegraphics[height=2.35cm]{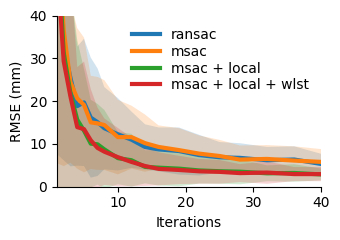}\label{fig:planarExtractionPlot}}\subfigure[Plate location]{\scriptsize{}\begin{tabular}[t]{cc}
			\includegraphics[width=0.32\columnwidth,trim={2.5cm 0cm 1.1cm 2cm},clip]{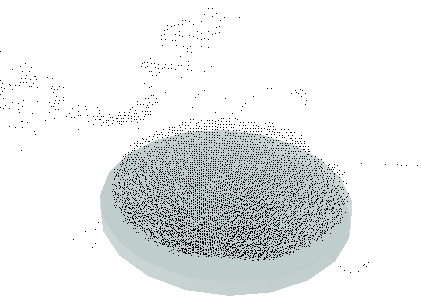}&\includegraphics[width=0.32\columnwidth,trim={2.5cm 0cm 1.1cm 2cm},clip]{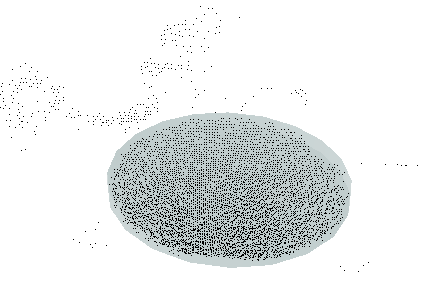}\\
			Naive&Projection weights (ours)
	\end{tabular}\label{fig:rpKDE}}
	
	\caption{Estimation of the turntable normal~\ref{fig:planarExtractionPlot} and location~\ref{fig:rpKDE}. \ref{fig:planarExtractionPlot} is the Root Mean Square Error (100 runs average) as the inliers distance to the ground truth plane. \ref{fig:rpKDE} shows the localization of the turntable with naive MeanShift~\cite{Comaniciu2002} or our projection weights. Using our weighed techniques, we can cope with heterogeneous data and estimate correctly the center.}
	\vspace{-1em}
\end{figure}

The normal $\vec{n}_i$ is estimated with an mSAC~\cite{Torr2000}, an improved RanSAC~\cite{Fischler1981}. Fig.~\ref{fig:planarExtractionPlot} benchmarks combinations of ransac/msac. Because SAC usually assumes noise-free models, we refine each estimation with local optimization (\textit{local}) and weighted least square (\textit{wlst}) as in~\cite{Chum2003,Lebeda2012}, which lead to better results. 
We used a standard probabilistic stopping criterion, assuming a weakly known ratio of object/scene points.

To estimate the location $\vec{c}_i$ of the turntable from the plane's inliers we use an improved version of MeanShift~\cite{Comaniciu2002} - a Kernel Density Estimator (KDE) - where a randomly-initialized kernel is iteratively shifted towards the center of gravity, until convergence (here, $3\text{mm}$). To cope with heterogeneous density data, we weight each point with its corresponding projected area on the sensor pixel given the normal $\vec{n}_i$ and pixel depth.
The intuition is to value less the high density points which are closer to the depth sensor but image in fact a smaller part of the scene. Fig.~\ref{fig:rpKDE} demonstrates our weighted MeanShift converges correctly whereas the naive approach fails.

\subsection{Point clouds registration}
\label{sec:PCLRegistration}
\begin{sloppypar}
To register each point cloud $\mathcal{P}_i$ we need to estimate the unique transformation matrix $\mathbf{M_i}$ so that all points clouds are expressed in a common reference frame.
With the detected turntables in each point cloud we can estimate $\mathbf{M_i}$ \textit{up to} a rotational extent around the revolutionary axis.
Indeed, due to its radial symmetry the rotation around the revolution axis cannot be extracted.

Formally, we decompose $\mathbf{M_i}$ as two $4\times{}4$ transformation matrices, with: $\mathbf{M_i} = \mathbf{U_i}\cdot{}\mathbf{V_i}$. Where $\mathbf{U_i}$ aligns the center and normal of the turntable, and $\mathbf{V_i}$ rotates around the revolution axis. 
$\mathbf{U_i}$ is estimated from the turntable parametric model of each point cloud, that is:
{$\mathbf{U_i} = \begin{pmatrix}
\mathbf{R}_i & \mathbf{t}_i \\
0 & 1
\end{pmatrix}$} with $\mathbf{R}_i$ a 3x3 rotation matrix around the vector $\vec{n}_i\times{}\vec{n}_r$ with angle $cos^{-1}(\frac{\vec{n}_i\cdot{}\vec{n}_r}{||\vec{n}_i||})$, $\mathbf{t}_i$ a 3x1 vector defined as $\mathbf{t}_i = \mathbf{R}_i\cdot{}(\vec{c}_r-\vec{c}_i)$. The arbitrary target position and orientation is defined as $\vec{c}_r = (0,0,0)$ and $\vec{n}_r = (0, 1, 0)$.

\end{sloppypar}

$\mathbf{V_i}$ is a 4x4 transformation with no translation that rotates around the revolution axis $\vec{n}_r$ with an angle~$\phi_i$ which we experimentally defines from the setup.
\\

\begin{sloppypar}
The merging of all registered point clouds is denoted $\mathcal{P}_r$, such that: {$\mathcal{P}_r = \mathcal{P}_1\cdot{}\mathbf{M_1} \cap ... \cap \mathcal{P}_n\cdot{}\mathbf{M_n}$}.
\end{sloppypar}

\subsection{Modeling of revolving pottery object}
\label{sec:ModelingRevolvingObject}

From point cloud $\mathcal{P}_r$ we now seek to model the revolving object using its radial symmetrical property.
The underlying idea is to compute the radial accumulation of all points around the revolution axis $\vec{n}_r$ and to extract the object profile from high radial density areas. We name this \textit{radial accumulator}.

\subsubsection{Building the radial accumulator}
\label{sec:buildradialacc}

\begin{figure}[t]
\centering
	\subfigure[Polar space]{\includegraphics[width=0.55\columnwidth]{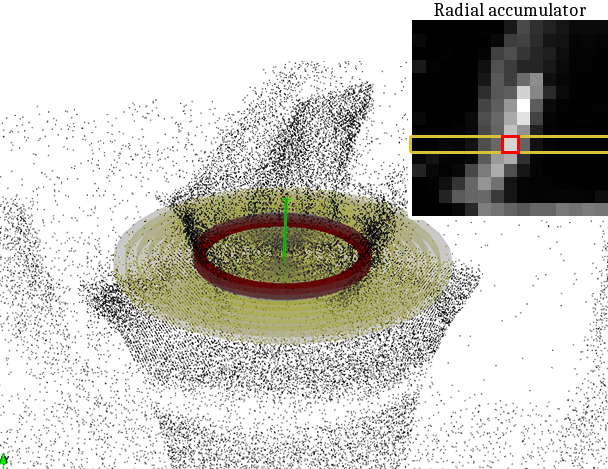}\label{fig:radAccAnnulii}}\hspace{0.03\linewidth}
	\subfigure[Heterogenous radial density]{\includesvg[width=0.41\columnwidth]{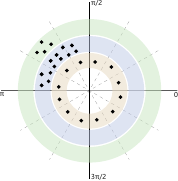}\label{fig:radAccUnevenDensity}}
	\caption{\ref{fig:radAccAnnulii}~Scene clustering in 3D annulii which densities are encoded in the radial accumulator (inset). The yellow annulii displayed correspond to the row highlighted in the accumulator. \ref{fig:radAccUnevenDensity}~2D Illustration of heterogeneous radial density. The blue and red annulii exhibit same density (ergo same accumulator value) but different radial distribution. We address this by weighting each accumulator cell with its radial spread.}
	\vspace{-1em}
\end{figure}

We first transform all the points into a polar coordinates $(\rho, h, \theta)$ where $(0,0,0)$ maps the turntable center $\vec{c}_r$, $\rho \in \mathbb{N}$ is the orthogonal-distance to $\vec{c}_r$, $h \in \mathbb{R}$ the elevation 
along the $\vec{n}_r$, and $\theta \in [0;2\pi[$ the angle of rotation.

If data were dense noise-free and holes-free, the profile of the object would be obtained by sampling all points having $h \geq 0$ within a small $\theta$ interval, thus providing a cross-sectional view of the object. In practice such approach fails since $\mathcal{P}_r$ is noisy, does not provide an all-around view of the object, and suffers from occlusion (i.e. Potter's hand). 
Instead, we accumulate data radially to reduce noise influence, and later account for radial spread to diminish the effect of occlusion.

To compute the radial accumulator we cluster the polar space into \textit{3D annulii} which volume is formed by a cylinder into which we subtract an other cylinder of smaller radius. Fig.~\ref{fig:radAccAnnulii} illustrates the 3D clustering into annulii.
For each annulus of location $(\rho, h)$ width $\Delta_{\rho}$ and height $\Delta_{h}$, the accumulator $\Gamma$ is the density of data within the annulus:
\begin{equation}
\label{eq:accOriginal}
\Gamma(\rho, h) = \frac{|\mathcal{A}|}{\mathcal{A}_{vol}}\,,
\end{equation}
where $|.|$ is the cardinality of $\mathcal{A}$, the set of points in the annulus:
\begin{equation}
\mathcal{A} = \{\chi| \chi\in\mathcal{P}_r, \; \forall \chi_{\rho} \in [\rho, \rho+\Delta_{\rho}[, \chi_{h} \in [h, h+\Delta_{h}[\}\,.
\end{equation}

\begin{figure}[t]
	\newcommand\potFrame[1]{\includegraphics[width=0.22\linewidth]{ressources/visus/output/T1_claude_b5_cam0-1_acc_equi0_homo0_#1.png}}
	\newcommand\potFrameImp[1]{\includegraphics[width=0.22\linewidth]{ressources/visus/output/T1_claude_b5_cam0-1_acc_equi1-linear_homo2_#1.png}}
	\newcommand\potFrameHL[1]{\includegraphics[width=0.22\linewidth]{ressources/visus/output/T1_claude_b5_cam0-1_acc_equi0_homo0_#1_hl.png}}
	\newcommand\potFrameImpHL[1]{\includegraphics[width=0.22\linewidth]{ressources/visus/output/T1_claude_b5_cam0-1_acc_equi1-linear_homo2_#1_hl.png}}
	
	\centering
	\scriptsize
	\setlength{\tabcolsep}{0.0025\linewidth}
	\renewcommand{\arraystretch}{0.75}
	\begin{tabular}{cccccccc}
		&\multirow{1}{*}[1.2cm]{\rotatebox{90}{Original}}&\potFrame{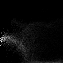}&\potFrameHL{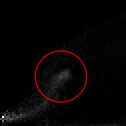}&\potFrameHL{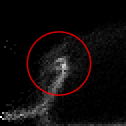}&\potFrameHL{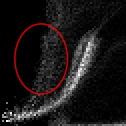}\\
\multirow{1}{*}[1.25cm]{\rotatebox{90}{Radially}}&\multirow{1}{*}[1.31cm]{\rotatebox{90}{enhanced}}&\potFrameImp{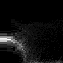}&\potFrameImpHL{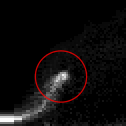}&\potFrameImpHL{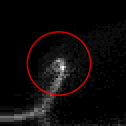}&\potFrameImpHL{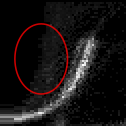}\\
&& (a) & (b) & (c) & (d) \\
	\end{tabular}

	\caption{Radial accumulator $64\times{}64$ (normalized for display), for chronologically ordered frames. Top shows original accumulator (eq.~\ref{eq:accOriginal}) and bottom shows radially enhanced accumulator (eq.~\ref{eq:accEnhanced}). Non-radial artifacts (red circle) such as the potter's hands are less visible in the enhanced version. Pictures are better seen on a screen.
	For all, $r=160\text{mm}$ and $\Delta_{h}=\Delta_{\rho}=\frac{r}{64}$.}
	\label{fig:radAccs}
	\vspace{-1em}
\end{figure}

Top row fig.~\ref{fig:radAccs} shows a few radial accumulators
from a unique pottery recording ($\Delta_{\rho} = \Delta_{h} = r / 64$). 
Despite noise and occlusions, the evolution of the object profile is clearly visible starting from a half-sphere shape (a) to a large bowl shape (d). 
Still, there are noticeable artifacts as the accumulator is affected by the presence of local non radial elements such as the potter's hand on/inside the bowl (a,d), or the fingers pinching the tip of the bowl (c).
To cope with this we use the data radial spread.

\begin{sloppypar}
\paragraph{Radial spread}

The underlying problem of the radial accumulator is that it is highly affected by non-radial high density data, which is illustrated in fig.~\ref{fig:radAccUnevenDensity} where the blue and red annulii have same density, ergo same accumulator value.

We address this issue by promoting annulii with data well radially distributed and subsequently weight the accumulator with the \textit{radial spread} of each annulus.
We borrowed a radial spread metrics from the field of circular statistics, which is defined as the length of the resultant vector $\bar{r}$~\cite{jammalamadaka2001topics}:
$\bar{r} = \frac{1}{|\mathcal{A}|}\sum_{i}r_i$, with
$r_i = \begin{pmatrix}
\cos \theta'_i \\
\sin \theta'_i
\end{pmatrix}$. Different from the original metric, for our use $\theta'_i$ is the $\theta_i$ polar coordinate of a point in $\mathcal{A}$ after being normalized with the min and max $\theta$ of the points in the annulus.
The purpose of doing that is to avoid penalizing setups with fewer sensors. 
Without normalization, annulii covering only a small cross-sectional view of the object would have a lower accumulator value leading to less accurate object reconstruction.

Subsequently, we redefine the accumulator with:
\begin{equation}
\Gamma(\rho, h) = (1-||\bar{r}(\mathcal{A})||) \times \frac{|\mathcal{A}|}{\mathcal{A}_{vol}}\,.
\label{eq:accEnhanced}
\end{equation}

\noindent{}Sample outputs with the radially enhanced accumulator are shown in the bottom row fig.~\ref{fig:radAccs}. As expected, it reduces artefacts which is especially visible in (d) where the potter's hand is significantly less visible in the enhanced accumulator.
\end{sloppypar}

The only parameter for this stage is the cell size ($\Delta_{\rho} = \Delta_h$) which is set given the targeted accumulator resolution. Its influence is analyzed in sec.~\ref{sec:expAccSize}.

\subsubsection{Profile extraction}
\label{sec:profileextraction}

We now use the radially redundant data from the accumulator to model the profile of the object. The task is challenging due to noisy data containing outliers and because pottery objects exhibit various shapes and evolve unpredictably with time.

\paragraph{Profile model}
A 3D object's profile is defined as a 2D parametric curve (possibly bijective) in the polar space. 
B-Splines are often used for such purposes~\cite{Wang2006a,Zheng2012} as they are piece-wise combination of polynomials segments defined by a compact set of knots. B-Splines have their intermediate knots \textit{off} the curve which is convenient for numerical optimization (e.g. convex hull computation) but not optimal for us since: a) knots can be located outside the accumulator boundaries, b) small temporal profile changes may translate to large knots displacements.

We instead use Catmull-Rom~\cite{Catmull1974}, a form of cubic Hermite spline. 
Unlike B-Splines, Catmull-Rom have knots \textit{on} the curve which allows bounded search space.
Furthermore, the curve tension $\tau \in [0;1]$ is easily parametrized. Following~\cite{Yuksel2011}, we set $\tau=1.0$ (a.k.a chordal Catmull-Rom) as it produces smooth curve without self intersection.
Hence, the object's profile is modeled as $C^k$ a Catmull-Rom with $k$ knots $\{\kappa_1, ..., \kappa_k\}$. The coordinates from knots $j$ to $j+1$ is computed with $\chi(p)$ ($p \in [0;1]$ the progression):
\begin{equation}
\chi(p) = 
\frac{1}{2}
\begin{pmatrix}
1 & p & p^2 & p^3
\end{pmatrix}
\begin{pmatrix}
0 		&	2 		& 	0 			& 	0 		\\
-\tau	& 	0		&	\tau		&	0 		\\
2\tau	& 	\tau-6	& 	-2(\tau-3)	&	-\tau	\\
-\tau	&	4-\tau	&	\tau-4		&	\tau	
\end{pmatrix}
\begin{pmatrix}
\kappa_{j-1}	\\
\kappa_{j}		\\
\kappa_{j+1}	\\
\kappa_{j+2}
\end{pmatrix}\,.
\label{eq:splineCatmullRom}
\end{equation}
Note that, first ($\kappa_1$) and last knot ($\kappa_5$) are referred as \textit{virtual knots}~\cite{Wang1998,Yuksel2011}\footnote{First and last knots only control the starting/ending tangent, as eq.~\ref{eq:splineCatmullRom} is defined only for $j \in ]1;k[$.}
and computed with the axis-reflection proposed in~\cite{Wang1998}. Formally:
$\kappa_1 = \kappa_2 - |\kappa_3 - \kappa_2|$ and
$\kappa_k = \kappa_{k-1} - |\kappa_{k-2} - \kappa_{k-1}|$. This further simplifies the optimization and reduce the search space.
In practice we use Catmull-Rom with 5 knots ($C^5$) as it is sufficient to match all pottery shapes, thus leaving us with 6 unknown parameters (3 non-virtual knots with 2D coordinates).

Finding the optimal matching $C^5$ curve can be done with non-linear fitting~\cite{Wang2006a} but would assume unrealistic noise-free data. To gain robustness to noise (outliers) and benefit of temporal information, we use a particle filter.

\paragraph{Particle Filter}
The particle filter implemented is a bootstrap~\cite{Candy2007}. Suppose a set of $N$ particles (curves) denoted $X \leftarrow \{C^5_1, ..., C^5_N\}$, after each frame we update these particles with a motion function $\lambda$ so $X \leftarrow \lambda(X)$ and our filter evaluates ${P(C^5_i|\Gamma) \forall i \in [1,N]}$ the probability of each particle ($C^5_i$) to match the current observation (the accumulator $\Gamma$).
After each run, we resample a new set of particles $X$ with a systematic resampling $f$~\cite{li2015resampling}, $X \leftarrow f(X)$. As opposed to the classical \textit{roulette} sampling, the systematic resampling better preserves particles with low probability which thus avoids overfitting.

To initialize the particle filter, we randomly draw $N$ particles $C^5$. 
Consider a particle $C^5_i$, its non-virtual knots $\kappa_{i,j}$ (with $j \in ]1;5[$) are initialized such that $\kappa_{i,j} = (U, U)$ where $U \in [0;r]$ is a uniform function ($r$ the turntable radius). 
As the object's profile evolves temporally, we account for the expected motion by updating particles knots with a random vector drawn from a 2D zero-centered symmetrical Gaussian distribution $S$:
\begin{equation}
\label{eq:particleUpdate}
\kappa_{i,j} \leftarrow \kappa_{i,j}+S(\sigma_m)\text{, with }j \in ]1, 5[\,.
\end{equation}
The $\sigma_m$ parameter reflects the changes along the $\rho$ and $h$ axes of the object's profile through time. Experimentally\footnote{This corresponds to a maximum motion of $50\text{mm}$ per second in $\rho$ and/or $h$ axes, which reflects the fast object shape motion in some modeling phases.}, $\sigma_m = 2mm$.\\

\begin{figure}[t]
\centering
	\includegraphics[width=0.4\columnwidth]{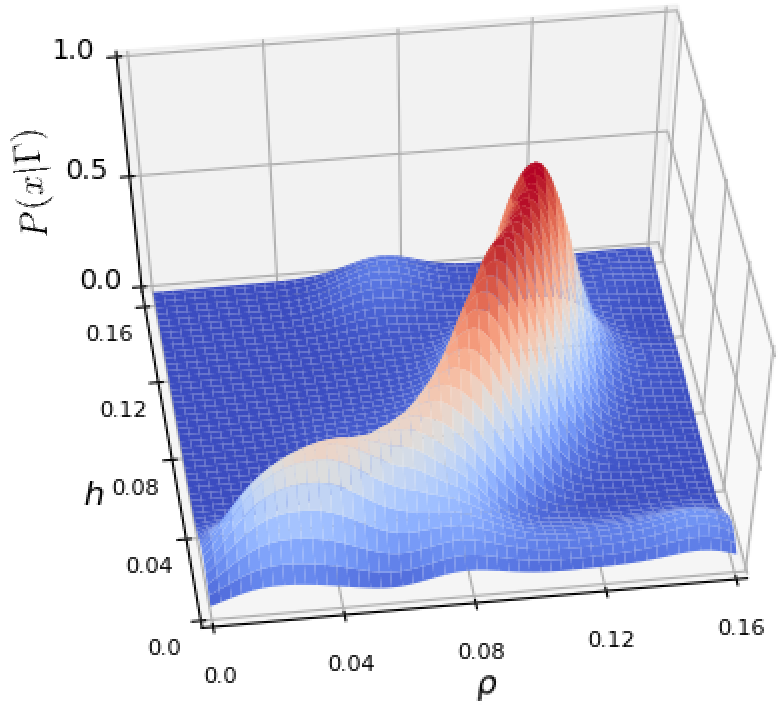}
	\caption{Radial accumulator represented as weighted Gaussian Mixture. $P(x|\Gamma)$ is the weighted average of the top 10 Gaussians (eq.~\ref{eq:dataScoring}).}
	\label{fig:gmm}
	\vspace{-1em}
\end{figure}
\begin{sloppypar}
The distinctive feature of our particle filter is the evaluation of a particle against the accumulator (scoring). 
We consider the accumulator $\Gamma$ as a mixture of $m$ weighted Gaussians $w\mathcal{N}$ such that the centers and weights of the Gaussians are the locations and values of the accumulator, respectively. 
Given $\phi(x | \mathcal{N})$, the Probability Density Function~(PDF) of a Gaussian, we estimate the likelihood of a particle to belong to the set of Gaussians. 
In a standard Gaussian Mixture Model~(GMM), the probability of $x$ to belong to a GMM is the maximum of all Gaussians PDF. 
Instead, we use the mean of the top 10 weighted Gaussians PDFs which better accounts for spatial relationship of the data and is found to be significantly more robust to noise.
\end{sloppypar}

\begin{sloppypar}
\noindent{}Consider a data $x$, the resulting set of weighted Gaussians PDF for the accumulator is denoted $\{w_0\phi_0(x), w_1\phi_1(x), ..., w_m\phi_m(x)\}$ and is sorted in descending order (i.e. $w_i\phi_i(x) \geq w_{i+1}\phi_{i+1}(x)$). The probability $P(x|\Gamma)$ is then obtained with:
\begin{equation}
P(x|\Gamma) = \frac{w_0\phi_0(x) + w_1\phi_1(x) + ... + w_9\phi_9(x)}{10}\,.
\label{eq:dataScoring}
\end{equation}
The weighted Gaussian Mixture is illustrated in fig.~\ref{fig:gmm} for a ball-shaped pottery object.
By extension, the probability of the particle $C^5_i$ to match the current accumulator is computed with:
\begin{equation}
P(C^5_i|\Gamma) = \frac{1}{||C^5_i||}\sum_{x \in C^5_i}^{} P(x|\Gamma)\,,
\label{eq:particleScoring}
\end{equation}
where $x \in C^5_i$ denotes points lying on the Catmull-Rom curve $C^5_i$, obtained through equidistant sampling.
The score is divided by the curve length $||C^5_i||$ to avoid favoring longer curves.
Note also that all Gaussians variance are arbitrarily defined as the accumulator cell size (i.e. $\Delta_{\rho}$). This allows to remain invariant to changes in the accumulator resolution, and also helps the particle filter to converge faster.
\end{sloppypar}

Because of the high search space dimension, averaging top particles has no sense so we simply use the $C^5$ particle with best score (eq.~\ref{eq:particleScoring}) as our object profile. 
The final 3D mesh is then computed through radial symmetry around revolution axis.

\begin{sloppypar}
\paragraph{Technical implementation}
\label{sec:aboutcomplexityandsearchspace}
To speed up the particles filter update, we use a vectorized implementation with an $O\big(N(m\log(m)) + N\log(N)\big)$ complexity ($N$ the number of particles, $m$ the number of accumulator cells) and constrain the first non-virtual knots ($\kappa_{2}$) on the $h$ axis to preserve revolving properties and reduce the search space.

\end{sloppypar}

\begin{sloppypar}
Our complete pipeline algorithm is detailed in~\ref{sec:algorithm}. 
\end{sloppypar}

\section{Experiments}

For evaluation, we acquired 3 pottery sequences (total 6030 frames) with two small and non-invasive depth sensors PMD nano camboard which output 160x120 depth maps @25FPS with $90^\circ$ field of view. 
Each sequence was complex to record as it shows a professional potter in its own artist studio during a complete making bowl process, starting with an empty turntable. 
Following the potter's gesture study~\cite{Manitsaris2013} the sensors were approximately mounted at shoulder's height to ensure that the main motion axis is not aligned with the camera (depth) axis which is the less precise.
Depth maps are converted to 3D points clouds using the intrinsic camera calibration parameters. 

\paragraph{Data labeling}
Frames were manually labeled by operators relying on the 3D scene view (manually registered point clouds). The operators were asked to label the object profile with as many points as needed, which we then fitted to $C^5$ Catmull-Rom with a greedy spline fitting algorithm.\\
\textbf{Data and annotation will be released upon publication}.

\begin{figure}[t]
	\centering
	\includegraphics[width=0.95\linewidth]{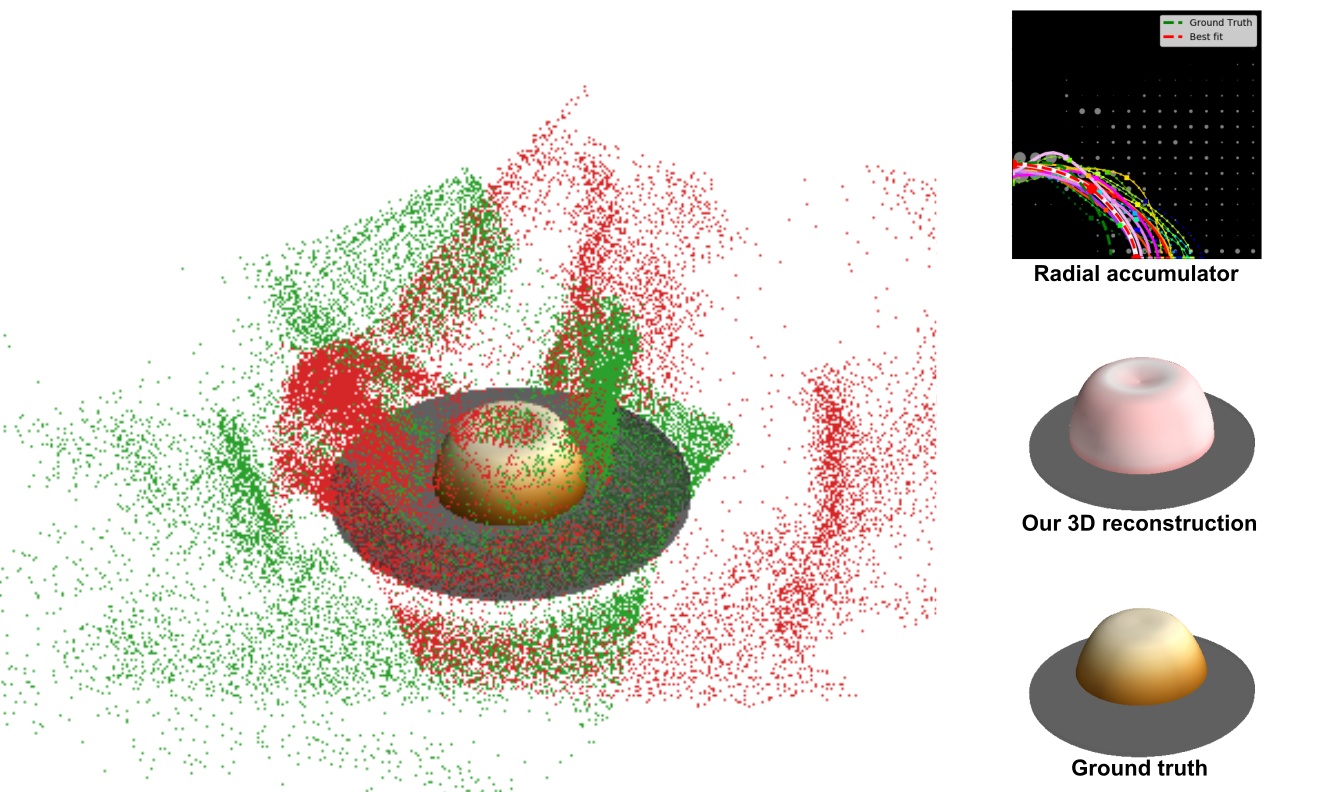}
	\caption{Detailed output of our method showing the 3D scene with point cloud registration, the detected turntable (gray cylinder), and our reconstructed 3D revolving object (height is color encoded). Right insets display from top to bottom, the radial accumulator with best particles, the 3D reconstruction of the object (cf. fig.~\ref{fig:outputObjectAdditional} for color scale) and the ground truth. The turntable is shrinked in the insets for visualization purposes.}
	\label{fig:outputObject}
	\scriptsize
	\setlength{\tabcolsep}{0.001\linewidth}
	\renewcommand{\arraystretch}{0.5}
	\begin{tabular}{cccccc}
		\multirow{1}{*}[1.6cm]{\rotatebox{90}{Reconstruction}\hspace{0.175cm}}&\includegraphics[width=0.035\columnwidth]{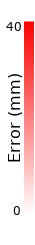}&\includegraphics[width=0.20\columnwidth]{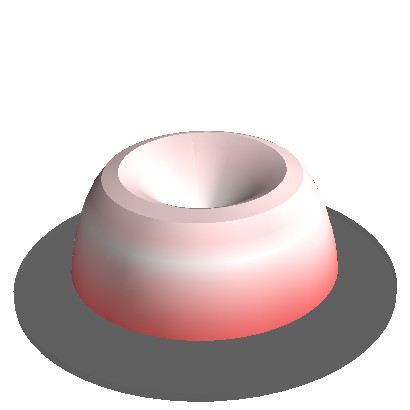}&\includegraphics[width=0.20\columnwidth]{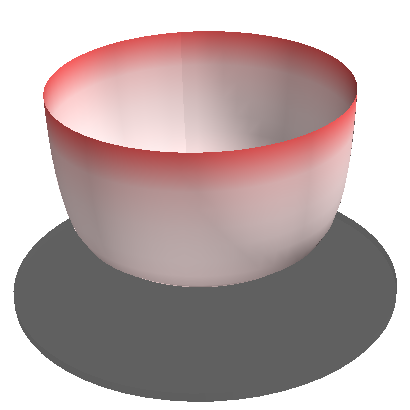}&\includegraphics[width=0.20\columnwidth]{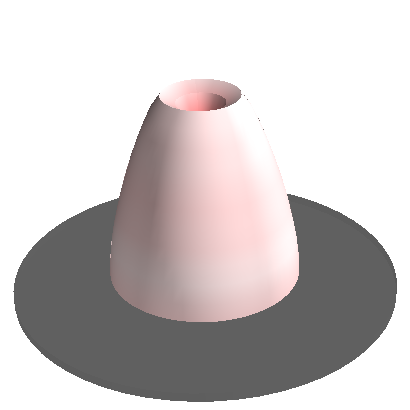}&\includegraphics[width=0.20\columnwidth]{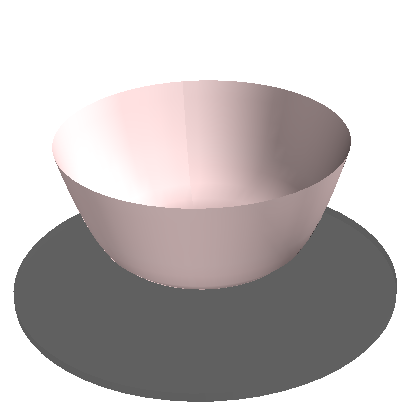}\\
		\multirow{1}{*}[1.55cm]{\rotatebox{90}{Ground truth}\hspace{0.15cm}}&\includegraphics[width=0.035\columnwidth]{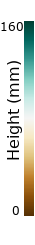}&\includegraphics[width=0.20\columnwidth]{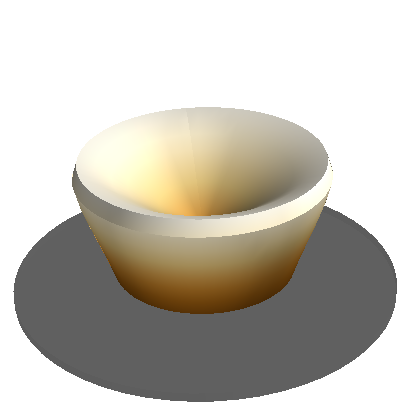}&\includegraphics[width=0.20\columnwidth]{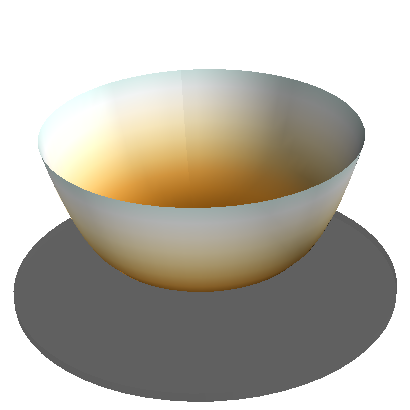}&\includegraphics[width=0.20\columnwidth]{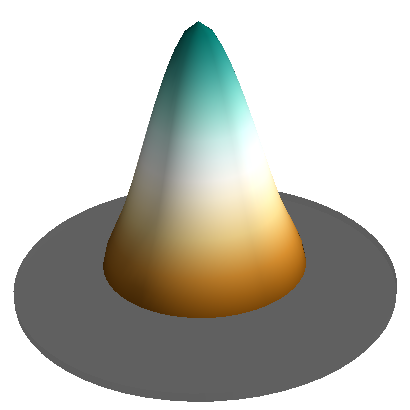}&\includegraphics[width=0.20\columnwidth]{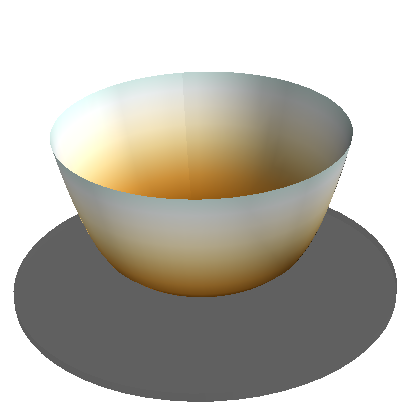}
	\end{tabular}
	
	\caption{Sample reconstruction with our method (top) and the corresponding ground truths (bottom). Top row: the color indicates the reconstruction error as the minimum local distance to the ground truth (from white=0mm to red=40mm). Bottom row: the color encodes the object height}
	\label{fig:outputObjectAdditional}
	\vspace{-2em}
\end{figure}

\begin{sloppypar}
\paragraph{Metrics for evaluation}
The prediction quality is measured with metrics derived from the shape matching literature~\cite{veltkamp2001shape}. 
The symmetrical Average Error reconstruction is defined as $\bar{\delta}_{AE}(P, \hat{P}) = \frac{\delta_{AE}(P, \hat P) + \delta_{AE}(\hat P, P)}{2}$, with $\delta_{AE}(A, B) = \text{avg}_{a \in A}\{\min{}_{b \in B}\{||a-b||\}\}$
where $||.||$ is the euclidean distance. $P$ and $\hat P$ are set of points equidistantly sampled every $0.2$mm from respectively, the ground truth profile and the predicted profile. 
The maximum reconstruction error is provided with the symmetrical Haussdorf Distance defined as $\bar{\delta}_{HD}(P, \hat{P}) = \frac{\delta_{HD}(P, \hat P) + \delta_{HD}(\hat P, P)}{2}$, with $\delta_{HD}(A, B) = \max{}_{a \in A}\{\min{}_{b \in B}\{||a-b||\}\}$.\\
Note that, using symmetrical (\textit{undirected}) errors the metrics reflect the completeness of the profile prediction. We also \textit{always} report 10 runs-average errors since particle filter is stochastic.
\end{sloppypar}

\subsection{Evaluation}
\begin{figure}[t]
	\centering
	\includegraphics[width=1.0\columnwidth]{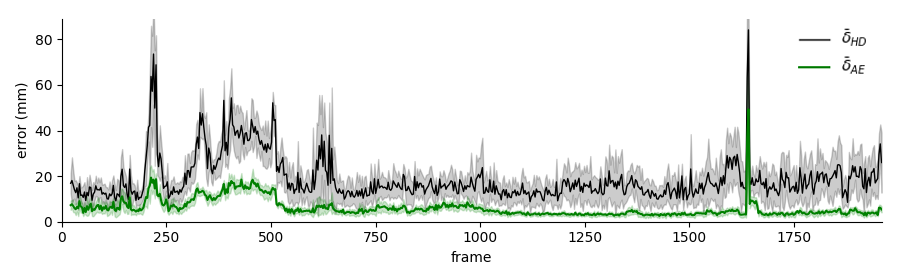}
	\vspace{-2em}
	\caption{Frame-wise errors (10-runs average) on a sample full bowl making sequence. $\bar{\delta}_{AE}$~(green) and $\bar{\delta}_{HD}$~(black). Note the error peaks (approx. frame 250 and 1700) reflecting large occlusion of the pottery object.}
	\label{fig:perfPlotSeq}
\end{figure}
\begin{sloppypar}
Tab.~\ref{tab:perfMethods} shows our performance on our full dataset using a 16x16 radial accumulator (i.e. $\Delta_{h}=\Delta_{\rho}=\frac{r}{16}=10mm$) and 1000 particles. 
We compare against results from the closest work in spirit, \cite{Wang2006a}, using a public implementation\footnote{https://github.com/Joey4s6l/BSplineFitting} and cherry picking their parameters with a grid search.
As expected, our method is significantly better than \cite{Wang2006a}. W.r.t. the ground truth, our reconstruction error is $\bar{\delta}_{AE}=7.60$mm and $\bar{\delta}_{HD}=19.84$mm, whereas \cite{Wang2006a} is at least twice worse.
This results of the B-Spline minimization used in \cite{Wang2006a} which subsequently implies no occlusion.
Despite the partially occluding potter's hands our method reconstructs efficiently the 3D object shape. 
Noticeably, our average error is smaller than the accumulator cell size which proves the benefit of the continuous Gaussian Mixture representation for particles evaluation. 
For fair comparison with \cite{Wang2006a}, we also report results without temporal filtering (\textit{Temp.} col in tab.~\ref{tab:perfMethods}) or using B-Spline with more knots.
Our method is better in all settings.
When temporal filtering is applied, we use a particles resampling ratio of $0.8$ which helps recovering from failures since $20\%$ of the particles are randomly drawn.
The relatively high variance of the metrics results of pottery occlusions as visible in sequence result fig.~\ref{fig:perfPlotSeq}, where error peaks correspond to major hand interactions.
\end{sloppypar}
A detailed visual output of our method is shown in fig.~\ref{fig:outputObject} and more compact outputs are in fig.~\ref{fig:outputObjectAdditional}.
Qualitatively, our method is able to reconstruct 3D revolving objects successfully. Despite the part-time presence of the potters' hands in or around the pottery (as in fig.~\ref{fig:outputObject}), our reconstruction is robust. Most noticeably the errors are located at the tip of the object (as in 2nd/3rd col fig.~\ref{fig:outputObjectAdditional}) which result of the sensor noise and the smooth Gaussian Mixture used to model the data. 
In some cases, potters also pinch the pottery object or fully cup the clay with their hands which affects the radial estimation (1st col, fig.~\ref{fig:outputObjectAdditional}).

We now study each parameter individually, without using temporal filtering.

\begin{table}[t]
	\scriptsize
	\centering
	\setlength{\tabcolsep}{0.01\linewidth}
	\renewcommand{\arraystretch}{1.0}
	\begin{tabular}{cccc}  
		\toprule
		& Temp. & $\bar{\delta}_{AE}$ (std.) & $\bar{\delta}_{HD}$ (std.) \\
		\midrule
		Ours & $\times$ & {8.09} (8.59) & {21.16} (15.56) \\
		Ours & \checkmark & \textbf{7.60} (8.64) & \textbf{19.84} (18.70) \\
		\midrule
		\cite{Wang2006a} (5 knots) & $\times$ & 16.08 (9.71) & 50.87 (20.92) \\
		\cite{Wang2006a} (8 knots) & $\times$ & 21.41 (8.08) & 74.71 (19.91) \\
		\bottomrule
	\end{tabular}
	\caption{Reconstruction errors (mm) on our dataset. \textit{Temp.} is temporal filtering ($0.8$ particles resampling). Our method exhibits significantly better performance for all setups.}
	\label{tab:perfMethods}
	\vspace{-1em}
\end{table}
\begin{figure}[t]
	\centering
	\subfigure[Sensors]{
		\setlength{\tabcolsep}{0.02\linewidth}
		\renewcommand{\arraystretch}{1.0}
		\scriptsize
		\begin{tabular}{ccc}
			\toprule
			Sensor & $\bar{\delta}_{AE}$ & $\bar{\delta}_{HD}$ \\
			\midrule
			1 & 9.77 & 25.76 \\
			2 & 10.56 & 23.56 \\
			1 \& 2 & \textbf{8.09} & \textbf{21.16} \\
			\bottomrule
		\end{tabular}\label{tab:perfSensors}
	}\hspace{0.2\linewidth}
	\subfigure[Number of particles]{
		\begin{tabular}{c}\includegraphics[width=0.35\columnwidth]{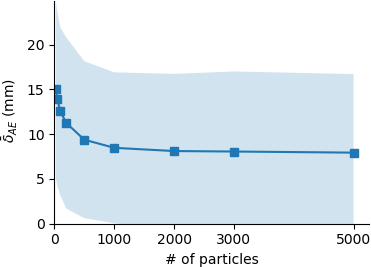}\end{tabular}\label{fig:perfParticlesN}}
	\vspace{-0.5em}
	\caption{Errors (mm) while varying sensor setups \ref{tab:perfSensors} or number of particles \ref{fig:perfParticlesN}. When varying sensors setup we used 1000 particles and use either sensor \textit{1}, \textit{2} or both \textit{1 \& 2}. When varying particles, both sensors are used.}
\end{figure}

\paragraph{Sensors ablation}
\begin{table}
\end{table}
To verify the robustness of our approach to the number of sensors input we evaluate our datasets using either sensor "1", sensor "2", or all sensors "1 \& 2" and report results in tab.~\ref{tab:perfSensors}.
With two sensors our method benefits of the point cloud registration and gets an $\bar{\delta}_{AE}$ which is $17.20\%$ better than with only the best of the two sensors (8.09mm vs 9.77mm).
Also expected, the maximum reconstruction error $\bar{\delta}_{HD}$ falls by $10.19\%$ using both sensors as compared to using only one sensor (21.16mm vs 23.56mm). 
This is explained by the larger radial field of view when using two sensors, and the relatively smaller impact of local noises.

\paragraph{Number of particles}
\begin{sloppypar}
Fig.~\ref{fig:perfParticlesN} shows that larger amount of particles improves reconstruction though the error quickly converges. 
In details, with 1000 particles the reconstruction is $19.74\%$ more precise than with 100 particles ($8.09$mm vs $10.07$mm), while 5000 particles is only $+4.60\%$ better than 1000 particles ($7.71$mm vs $8.09$mm). 
Noteworthy, relative to the time-complexity for 1000 particles the processing time for 200/5000/25000 particles is x0.5/x3.9/x17.9 slower. 
We argue that 1000 particles is a good performance/processing trade-off.
\end{sloppypar}

\paragraph{Resolution of the accumulator}
\label{sec:expAccSize}

\begin{figure}[t]
	\tiny{}
	\centering
	\subfigure[Sample outputs]{
		\scriptsize
		\setlength{\tabcolsep}{0.002\linewidth}
		\renewcommand{\arraystretch}{0.7}
		\begin{tabular}{cccc}
			\includegraphics[width=0.15\linewidth]{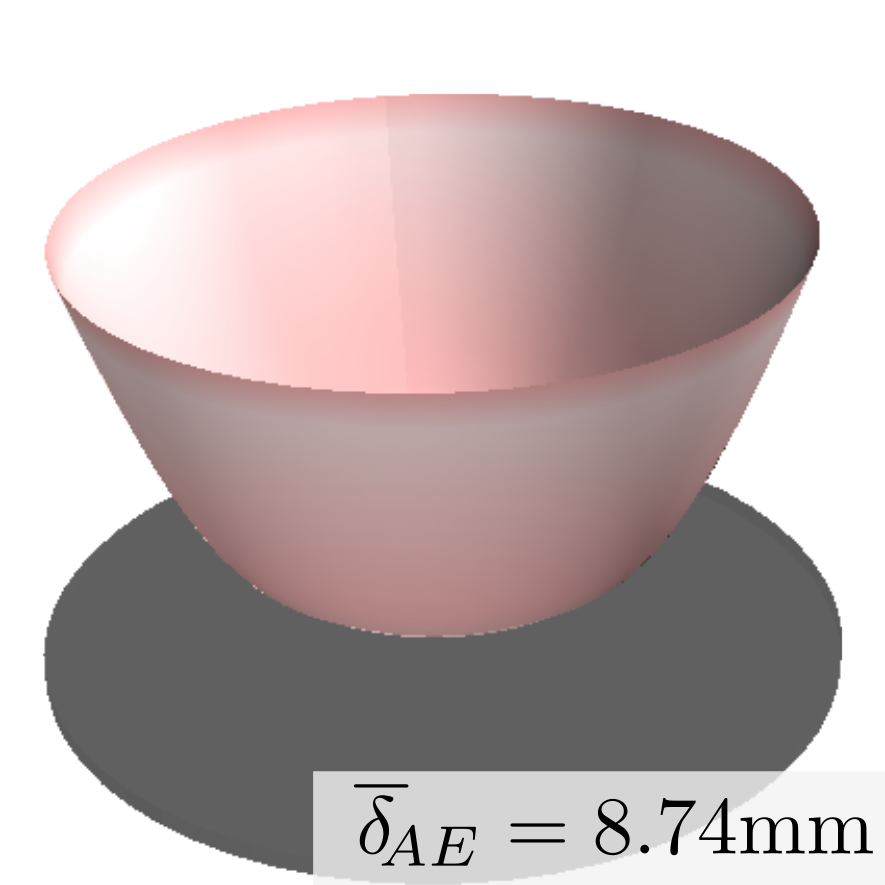}&\includegraphics[width=0.15\linewidth]{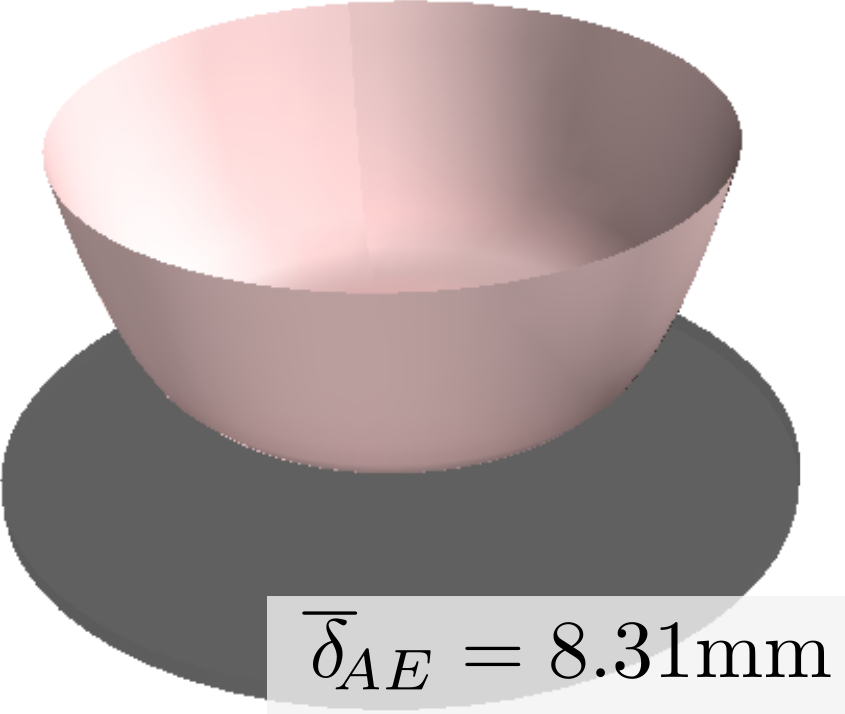}&\includegraphics[width=0.15\linewidth]{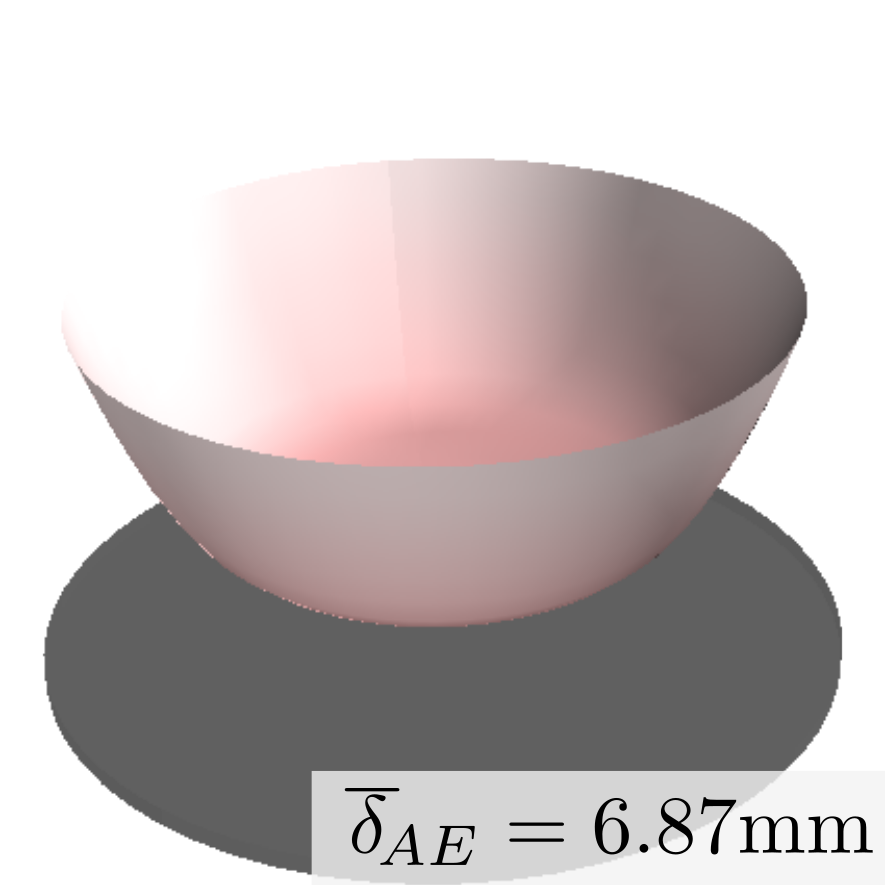}&\includegraphics[width=0.15\linewidth]{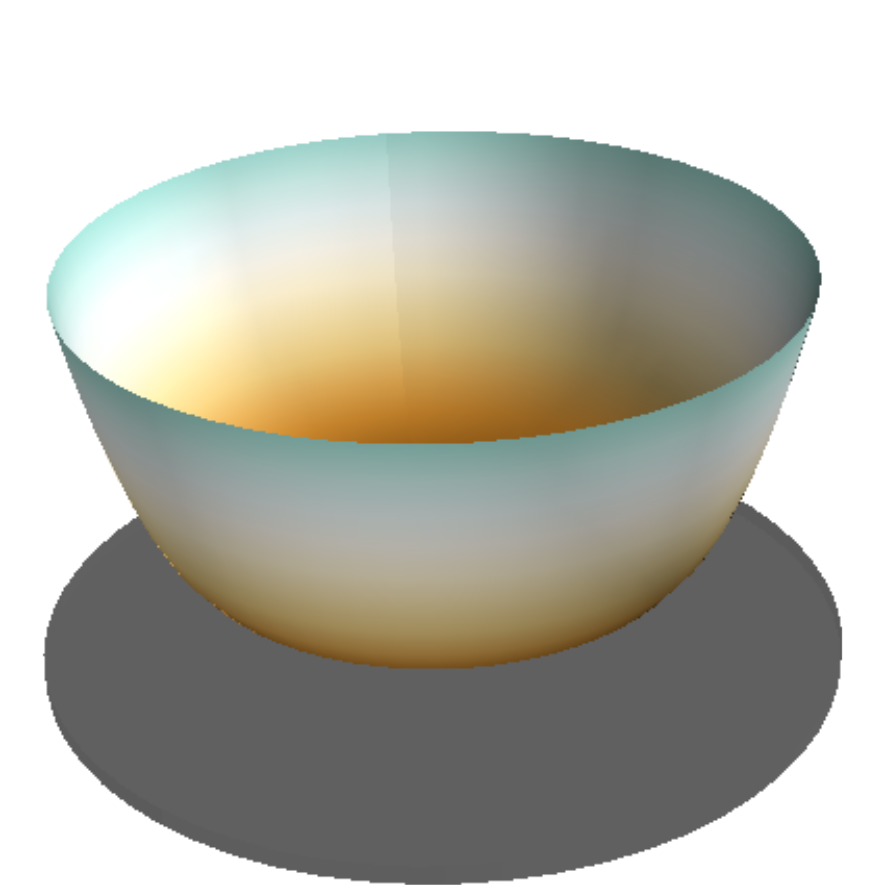}\\
			\\
			16x16 & 32x32 & 64x64 & Groundtruth
		\end{tabular}
	}\hspace{0.05cm}\subfigure[Performance]{
		\scriptsize
		\setlength{\tabcolsep}{0.018\linewidth}
		\renewcommand{\arraystretch}{1.0}
		\begin{tabular}{ccc}  
			\toprule
			Acc. size & $\bar{\delta}_{AE}$ & $\bar{\delta}_{HD}$ \\
			\midrule
			16x16 & 9.39 & 27.79 \\
			32x32 & 8.83 & 27.32 \\
			64x64 & \textbf{8.43} & \textbf{26.75} \\
			\bottomrule
		\end{tabular}
		\label{tab:perfAccSize}
	}
	\vspace{-1em}
	\caption{Study of the accumulator sizes. (a) Sample reconstruction show that reconstruction gets more precision with increasing resolution, which is also quantitatively proved with performance (mm) in (b).}
	\label{fig:perfAccSize}
	\vspace{-2em}
\end{figure}

\begin{sloppypar}
Intuitively, the resolution of the radial accumulator should affect the global reconstruction performance. 
The reason is that it discretizes the data in the $(\rho, h)$ polar space. 
Mean error metrics are reported in tab.~\ref{tab:perfAccSize} for accumulator sizes 16x16, 32x32 and 64x64 which correspond to cell sizes of $10$mm, $5$mm and $1$mm. 
As expected the average error $\bar{\delta}_{AE}$ decreases by $11.39\%$ for 64x64 accumulators compared to 16x16 ($8.43$ vs $9.39$) though it comes at the expenses of longer processing time. 
Measurements show that the processing time for accumulators of 32x32/64x64 is x2.2/x5.35 times slower than for 16x16. The effect of increasing resolution is visible from left to right in fig.~\ref{fig:perfAccSize}. Though with 16x16 accumulator the recovered shape is roughly correct, higher resolutions allow better profile modeling including more accurate detection of the tip of the object.
However, increasing accumulator resolution barely improves the average maximum error $\bar{\delta}_{HD}$ in tab.~\ref{tab:perfAccSize} proving that the large occlusions of the pottery object still affect the reconstruction.
\\
\end{sloppypar}

\begin{sloppypar}
Overall the proposed method reaches millimeters reconstruction error using one or more sensors. It also shows a great robustness to partial occlusion and great adaptation through time to the evolution of the object.
With large occlusions (e.g. hand covering the object) the problem however becomes ill-posed and our method cannot reconstruct the object properly. 
We argue that spatio-temporal deformation priors could be learned to handle complete occlusion during a small time period.
\end{sloppypar}

\section{Conclusion}
\label{conclu}

\begin{sloppypar}
We address 3D reconstruction of revolutionary object deformable through space and time.
Our method works with unconstrained setup providing one or more point clouds and reconstructs objects with $7.60$mm average error despite heavy hand interaction. 
Compared to the literature we are twice more precise in both mean and maximum errors, and our evaluation shows robustness to partial occlusion and sensors ablation. 

Future works will focus on estimating 3D gaussian mixture representation directly from point clouds.
\end{sloppypar}
\section*{Acknowledgements}
This work was funded by the European Commission via the iTreasures project (Intangible Treasures - Capturing the Intangible Cultural Heritage and Learning the Rare Know-How of Living Human Treasures FP7-ICT-2011-9-600676-iTreasures).

\bibliographystyle{model2-names}  %
\bibliography{library}

\appendix

\pagebreak
\section{Algorithm}
\label{sec:algorithm}

Algorithm \ref{alg:method} is the pseudo code of our entire method for modeling 3D revolving object, accounting for $n$ input sensors providing multiple point cloud inputs $\{\mathcal{P}_1, ..., \mathcal{P}_n\}$.
\makeatletter
\def\BState{\State\hskip-\ALG@thistlm}
\makeatother

\begin{algorithm}
\caption{Reconstruction of Revolving Objects}\label{alg:method}
\begin{algorithmic}[1]
	\State Initialize data storage
\algstore{bkbreak}
\end{algorithmic}\vspace{-0.75\baselineskip}
\rule{\columnwidth}{0.4pt}\vspace{-0.75\baselineskip}\flushleft{\textbf{Modeling of the turntable} (sec. \ref{sec:turntable})\vspace{-.01\textheight}}
\rule{\columnwidth}{0.4pt}
\begin{algorithmic}[1]
\algrestore{bkbreak}
	\Repeat
		\State{Read inputs $\{\mathcal{P}_1, ..., \mathcal{P}_n\}$}
		\For{$i\gets 1, n$}  %
			\State$\text{Estimate plane normal }\vec{n}_{i}^{t}\text{ in }\mathcal{P}_i$
			\State$\text{Estimate disk pos }\vec{c}_{i}^{t}\text{ in plane inliers}$
			\If{$|\vec{n}_{i}^{t}-\vec{n}_{i}^{t-1}| \leq \sigma_c$}
				\State Turntable $i$ is detected
			\EndIf
		\EndFor
		\State Next data frame
	\Until{All turntables are detected}
	\algstore{bkbreak}
\end{algorithmic}\vspace{-0.75\baselineskip}
\rule{\columnwidth}{0.4pt}\vspace{-0.75\baselineskip}\flushleft{\textbf{Point cloud registration} (sec. \ref{sec:PCLRegistration})\vspace{-.01\textheight}}
\rule{\columnwidth}{0.4pt}
\begin{algorithmic}[1]
\algrestore{bkbreak}
	\For{$i\gets 1, n$}
		\State$\text{Compute registration matrix }\mathbf{M_i}$
	\EndFor
\algstore{bkbreak}
\end{algorithmic}\vspace{-0.75\baselineskip}
\rule{\columnwidth}{0.4pt}\vspace{-0.75\baselineskip}\flushleft{\textbf{Modeling of revolving object} (sec. \ref{sec:ModelingRevolvingObject})\vspace{-.01\textheight}}
\rule{\columnwidth}{0.4pt}
\begin{algorithmic}[1]
\algrestore{bkbreak}
	\State Initialize accumulator $\Gamma$
	\State Initialize particles $\{C^5_1, ..., C^5_N\}$ randomly
	\Repeat
		\State{Read inputs $\{\mathcal{P}_1, ..., \mathcal{P}_n\}$}
		\State $\text{Register point clouds: }\mathcal{P}_{r} \gets \mathcal{P}_1\cdot{}\mathbf{M_1} \cap ... \cap \mathcal{P}_n\cdot{}\mathbf{M_n}$
		\State Update radial accumulator $\Gamma$ from $\mathcal{P}_{r}$ (sec. \ref{sec:buildradialacc})
		
		\For{$i\gets 1, N$} (sec. \ref{sec:profileextraction})
			\State Apply knots constraints on $C^5_i$
			\State Compute virtual-knots of $C^5_i$%
			\State Compute particle score $P(C^5_i|\Gamma)$%
		\EndFor
		
		\State Object profile is particle $C^5_{best}$ with highest score%
		\State Next data frame
	\Until{sequence ends}
\end{algorithmic}
\end{algorithm}

\end{document}